\documentclass[10pt,twocolumn,letterpaper]{article}

\usepackage{btas}
\usepackage{times}
\usepackage{epsfig}
\usepackage{graphicx}
\usepackage{amsmath}
\usepackage{amssymb}

\usepackage[pagebackref=true,breaklinks=true,letterpaper=true,colorlinks,bookmarks=false]{hyperref}

\btasfinalcopy 

\ifbtasfinal\fi

\let\OLDthebibliography\thebibliography
\renewcommand\thebibliography[1]{
  \OLDthebibliography{#1}
  \setlength{\parskip}{0pt}
  \setlength{\itemsep}{0pt plus 0.3ex}
}

\begin{document}

\title{SREFI: Synthesis of Realistic Example Face Images\thanks{* denotes equal contribution}}

\author{\parbox{16cm}{\centering
    {\large Sandipan Banerjee*$^1$, John S. Bernhard, Jr.*$^2$, Walter J. Scheirer$^1$, Kevin W. Bowyer$^1$,\\ Patrick J. Flynn$^1$}\\
    {\normalsize
    $^1$ Dept. of Computer Science \& Engineering, University of Notre Dame, USA\\
    $^2$ FaceTec, Inc.\\
        \tt\small \{sbanerj1, wscheire, kwb, flynn\}@nd.edu\\
        \tt\small jsbernhardjr@gmail.com
    }}
}

\renewcommand\footnotemark{}
\renewcommand\footnoterule{}

\maketitle

\begin{abstract}
In this paper, we propose a novel face synthesis approach that can generate an arbitrarily large number of synthetic images of both real and synthetic identities. Thus a face image dataset can be expanded in terms of the number of identities represented and the number of images per identity using this approach, without the identity-labeling and privacy complications that come from downloading images from the web. To measure the visual fidelity and uniqueness of the synthetic face images and identities, we conducted face matching experiments with both human participants and a CNN pre-trained on a dataset of 2.6M real face images. To evaluate the stability of these synthetic faces, we trained a CNN model with an augmented dataset containing close to 200,000 synthetic faces. We used a snapshot of this trained CNN to recognize extremely challenging frontal (real) face images. Experiments showed training with the augmented faces boosted the face recognition performance of the CNN.
\end{abstract}

\section{Introduction}

Researchers have assembled and shared face image datasets downloaded from the web, ranging from thousands to millions of images \cite{MegaFace,LFW,CASIA}. The VGG-Face dataset (2.6M images) \cite{VGG} is notable among these. However, industrial giants like Facebook \cite{Facebook_Deepface} and Google \cite{Google_FaceNet} have private datasets with over 200M face images at their disposal. It is difficult to compile such large datasets by selectively downloading images from the internet given the tremendous resource overhead.



\begin{figure}[t]
\begin{center}
   \includegraphics[width=1.0\linewidth]{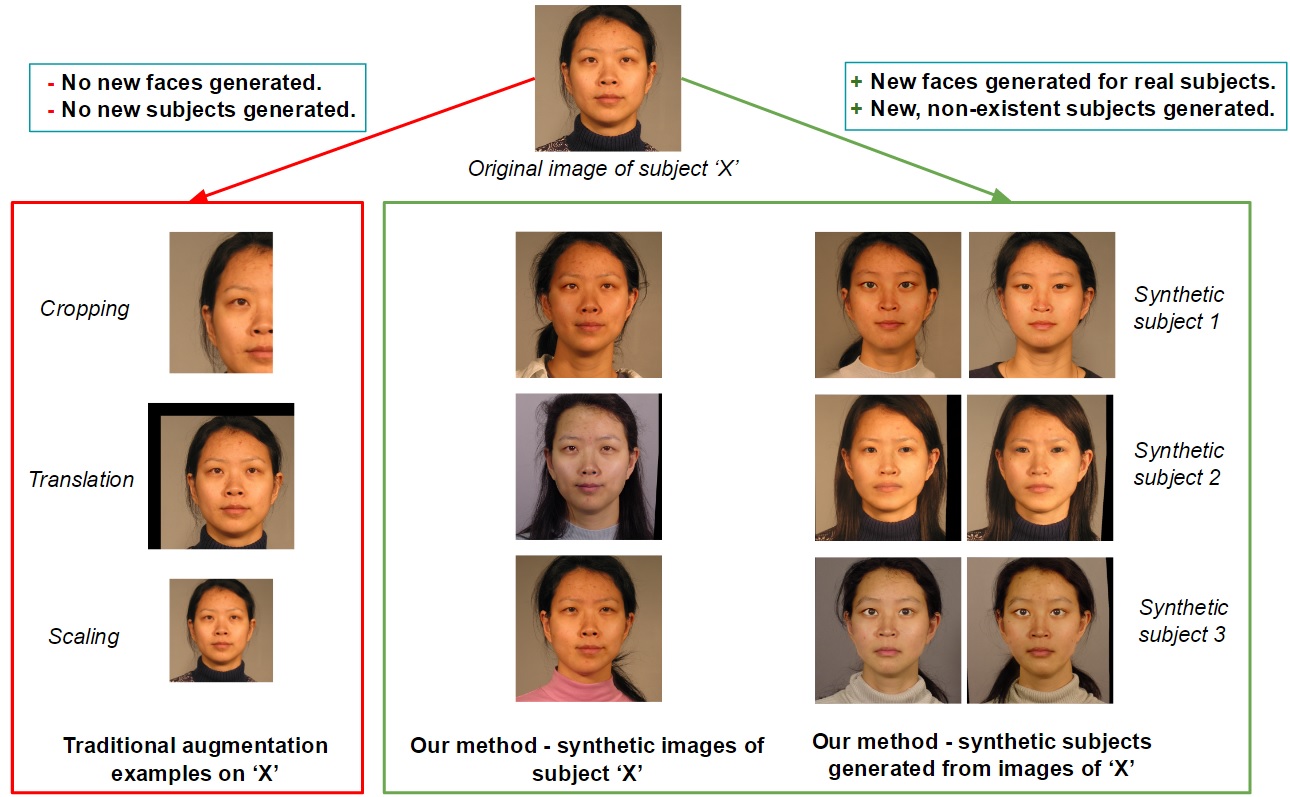}
\end{center}
   \caption{The motivation behind our work: while traditional augmentation has explored methods like cropping, translation, scaling \cite{AlexNet} (left column), we can augment a dataset by creating synthetic face images of existing identities (middle column) and new identities altogether (right column).}
\label{fig:SREFIMotivation}
\end{figure}

Recent results motivating the need for even larger datasets involve data augmentation for CNN training \cite{MasiAug}. It is common practice to augment any dataset, using simple methods like translation, cropping, rescaling, PCA whitening, etc., before using it to train a CNN to reduce the chance of it over-fitting on the training data \cite{AlexNet}. Despite these augmentation methods, CNNs tend to over-fit the training data, if each training class does not contain a considerable number of samples \cite{CNNSample}. For publicly available face datasets, often there are a few persons with a large number of images each, and a large number of persons with few images each. This is termed the `long-tail' problem in \cite{CASIA,MasiAug}. One result of this is that the CNN may fail to learn a feature representation that adequately represents the subjects with few images. Consequently, a CNN trained on such an unbalanced dataset might not be as effective as a CNN trained on a well-balanced dataset in face recognition tasks \cite{Google_FaceNet}.


There is a need to be able to create face image datasets that - (1) contain an arbitrarily large number of persons, (2) have a balanced number of images per person, and (3) do not run into potential issues of invasion of privacy. As a potential solution, we propose a system for automatically generating synthetic face images of either a real identity or a synthetic identity not corresponding to any real person. Combining these two modes, our approach can augment an existing dataset by increasing the number of face images per real identity, and by increasing the total number of (real + synthetic) identities, avoiding the long-tail problem. This is a significant improvement over traditional face augmentation methods (Fig \ref{fig:SREFIMotivation}). Also, when face images are generated for synthetic identities rather than any real person, privacy issues and the computation and communication burden of downloading from the web are avoided.


Our approach, ``SREFI", constructs synthetic face images, $512\times512$ in resolution, from a set of face images of real identities. The process starts with a real face image, the ``base face", divided into various region-specific triangles. A synthetic image is generated by stitching together corresponding triangles from other images (``donors") proximal to the base face in a CNN feature representation space. To obtain new synthetic images of a real identity, the donor images are selected from that real identity. To obtain images of a synthetic identity, the donor images are selected from a set of different real identities. Our approach re-shapes and adjusts the color distribution of the donor triangles, and blends them together to generate a natural-looking synthetic face image.

To validate our approach, we performed two types of face-matching experiments, on datasets augmented using this technique. One used human evaluators and the other used a pre-trained snapshot of the VGG-FACE model \cite{VGG}. These experiments show low intra-class variance between different images of a synthetic identity, and high inter-class variance between different synthetic identities. We performed another experiment to demonstrate the value of our synthetic image datasets in improving face recognition accuracy. We trained two networks with the VGG-FACE architecture from scratch - one on a batch of over 200,000 combined real and synthetic face images and the other with over 260,000 real face images from the CASIA-WebFace (CW) dataset \cite{CASIA}. We used these trained networks to recognize face images from the most challenging (`Ugly') partition of the `Good, Bad and Ugly’ (GBU) dataset \cite{GBU}. The CNN trained with the synthetic face images outperformed the CNN trained with the CW face images.


The remainder of this paper is structured as follows. Prior related research is discussed in Section 2. Section 3 describes the face images used as the donor set by our method. The synthesis pipeline is explained in detail in Section 4. Section 5 presents experimental results that assess the realism, uniqueness and stability of the synthetically generated face images and identities. Section 6 concludes the paper.

\section{Related Work}
Previous research related to this work can be categorized into three broad groups as below.

{\bf Face Recognition} - Face recognition performance is improving rapidly with the proliferation of data and deep CNN models. State-of-the-art face recognition algorithms have achieved near-perfect accuracy on the once-challenging LFW dataset \cite{LFW}. While some research focuses on creating novel CNN architectures \cite{VGG,DeepID3}, others focus on feeding a large pool of training data \cite{Facebook_Deepface,Google_FaceNet}. However, LFW with its 13,000 images is an under-representation of the diverse population of faces that a human encounters. The effect of this wide gap in data size on the recognition performance of the above methods has been documented in \cite{MegaFace}. Most of the recognition methods which produced near-perfect results on LFW (13k images) performed poorly in verification tasks on the MegaFace dataset (over 1M images) \cite{MegaFace}.

{\bf Face Synthesis} - One of the earliest face synthesis methods \cite{ICME05} used a combination of neighborhood patches from a set of images for hallucinating new faces. A more robust method was proposed in \cite{Bitouk08}, where the authors automatically selected a group of face images similar in appearance and pose to a source image. Each face in that group was overlaid on top of the source image and the combination with the lowest deviation from the original image was blended to give the final result. Others have implemented a model-based learning approach for different applications like swapping of 2D-aligned faces, expression flow across images of the same person and hallucination of new faces \cite{ICCV09,SIGGRAPH11,SIGGRAPH09,MM12,MIPRO14}. An inverse rendering approach for synthesizing a 3D structure of the face was proposed in \cite{PAMI13}, and more recently a one-shot version of the same was described in \cite{CogSci15}. The method proposed in \cite{ACCV14}, which divides the source face into four different parts and replaces each part from a gallery image before blending, is thematically closer to our approach. Recently, researchers have used generative adversarial nets (GANs) \cite{GAN} to change facial attributes like hair, mouth, eyeglasses or for artificially aging a face. However, GANs require a lot of training data and the synthetic face images are relatively low resolution and do not appear very realistic.

{\bf Data Augmentation} - Data augmentation has been known to help the learning of CNN models as it reduces overfitting on the training data \cite{DataAug}. Popular augmentation methods are simply over-sampling, mirroring or PCA whitening the training images\cite{AlexNet}. For re-identification purposes, another mode of augmentation proposed was to change the image background \cite{ReIDAug}. Hassner \etal proposed an augmentation method for faces \cite{Hass1,Hass2} which generates frontal versions of the face images to reduce recognition errors. This idea was extended in \cite{MasiAug} by augmenting the CASIA-WebFace dataset \cite{CASIA}, synthesizing new images with multiple pose, shape and expressions. However, a major difference between our method and these methods is that our system has the capability of increasing the number of training classes, by synthesizing new identities, while maintaining the number of images per subject.

\section{Collection of the SREFI Donor Set}
\begin{figure}[t]
\begin{center}
   \includegraphics[width=1.0\linewidth]{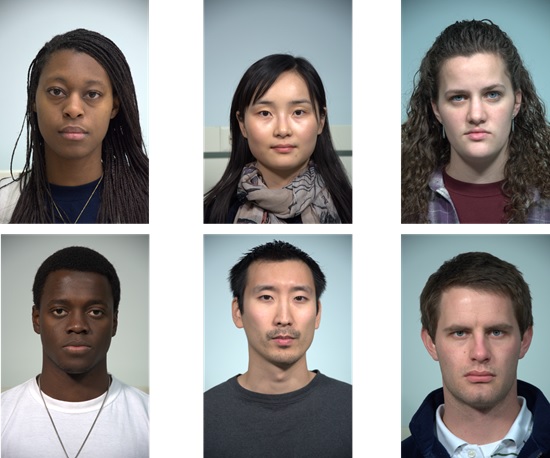}
\end{center}
   \caption{Example set of six images from the donor set.  A male and female of each race in the donor set are shown.}
\label{fig:DonorSetDist}
\end{figure}

The donor set for SREFI could be based on any dataset of real face images of a sufficient number of different persons. We used a dataset of donor face images created from an existing publicly-available dataset \cite{SREFIDonor}, in which multiple frontal images of each subject were acquired in different sessions. Subjects varied in gender, age and ethnicity. We used cropped, 2D-aligned and resized (to 512x512) versions of the images for our experiments (Fig \ref{fig:DonorSetDist}). For the sake of generating realistic synthetic face images, the donor set was subdivided by race and gender and images with very thick facial hair or glasses were removed. Table \ref{Tab:SREFIData} shows the image count and subject breakdown of the donor set.


\begin{table}
\caption{SREFI donor set distribution}
\vspace{0.1in}
\begin{small}
\begin{center}
\begin{tabular}{  |c | c| c|  }
\hline
Ethnicity & \begin{tabular}[x]{@{}c@{}}Male images\\(subjects)\end{tabular} & \begin{tabular}[x]{@{}c@{}}Female images\\(subjects)\end{tabular} \\
\hline
\hline
  Caucasian  &  7108 (678)  &  5510 (600) \\
  \hline
  Asian & 1903 (100)  & 1286 (74) \\
  \hline
  African American  &  67 (9)  &  65 (10) \\
\hline
\end{tabular}
\label{Tab:SREFIData}
\end{center}
\end{small}
\end{table}

\vspace{-0.1cm}
\section{Our Synthesis Method}
\vspace{-0.1cm}
The steps in our synthesis method, the workflow of which is depicted in Fig \ref{fig:SREFIWorkFlow}, are described below.


\subsection{Landmarking \& Triangulation}
\begin{figure*}
\begin{center}
   \includegraphics[width=1.0\linewidth]{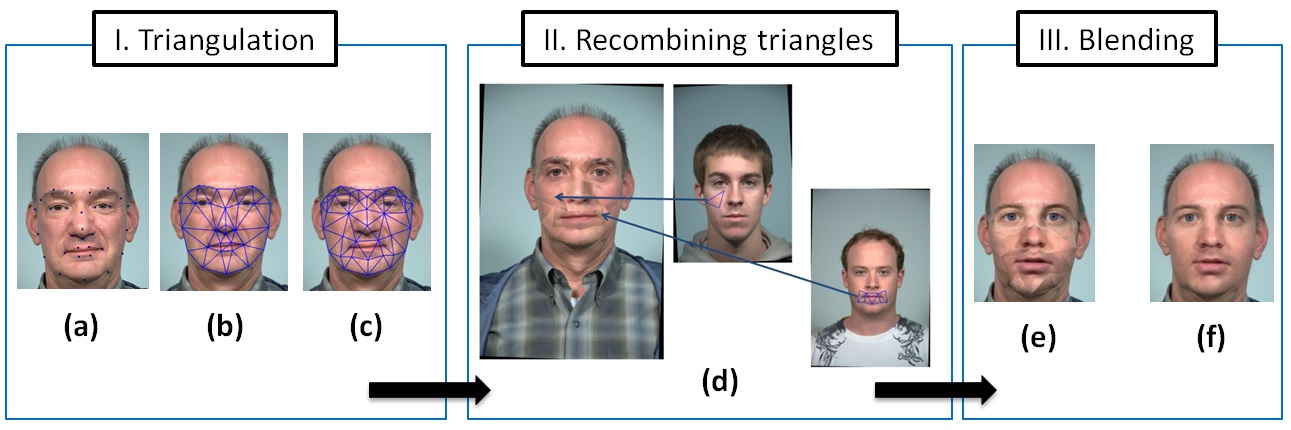}
\end{center}
   \caption{Workflow of the SREFI method on a high level.}
\label{fig:SREFIWorkFlow}
\end{figure*}

The base face was initially divided into triangular regions using a subset of the 68 facial landmark points acquired using the method from \cite{DLibPose} (Fig \ref{fig:SREFIWorkFlow}.a). Since these landmark points occur at important locations of the face, like the mouth, inaccuracies in detecting them would lead to artifacts (multiple mouths) appearing in the synthesized faces. To address this, we developed a triangulation that moved important facial features away from triangle corners, resulting in more visually stable regions. Using the landmark points and our initial triangulation of the face (Fig \ref{fig:SREFIWorkFlow}.b), we obtained centroids using the three vertices of a triangle. We created a new triangulation of the face by joining a centroid with the adjoining centroids of the triangles in its neighborhood (Fig \ref{fig:SREFIWorkFlow}.c). This allowed each region to be replaced in the target face using the region shape from the donors. Additionally, the outer part of the image was triangulated in order to allow the outer shape of the face to be modified. Our triangulation method is different from the one proposed in \cite{ACCV14}, as they used pre-defined points from the barycentric averages of landmark points extracted using the method from \cite{ZR}, to create masks of only three main facial regions.



\subsection{Selection of the Donor Pool}
We constructed the donor pool by selecting proximal face images to the base face in a lower-dimensional feature space. For this purpose, we extracted the 4096-dimensional feature vector for each face image in our donor set (Table \ref{Tab:SREFIData}) from the fc7 layer of a snapshot of the VGG-FACE CNN model \cite{VGG} pre-trained on 2.6M face images in the VGG-FACE dataset \cite{VGG}\footnote{Available here: \url{https://github.com/BVLC/caffe/wiki/Model-Zoo}}. We computed the mean feature vector for each subject of the dataset by averaging the feature vectors for each image of that subject.

The similarity score between two images was calculated using these features. Any distance function can be used for scoring: we chose cosine similarity as it has been extensively used to match VGG-FACE features by researchers \cite{PaSC}. The similarity score $S(v_{1},v_{2})$ between feature vectors $v_{1}$ and $v_{2}$ was calculated as:-
\begin{equation}
S(v_{1},v_{2}) = \frac{v_{1} \cdot v_{2}}{\left \| v_{1} \right \|_{2}\left \| v_{2} \right \|_{2}}
\end{equation}

where $S(v_{1},v_{2})$ is the similarity score between two vectors $v_{1}$ = [$f_1$, $f_2$, $f_3$, ..., $f_{4096}$] and $v_{2}$ = [$f_1$, $f_2$, $f_3$, ..., $f_{4096}$]. If $v_{1}$ and $v_{2}$ are perfectly alike (0 angle), then $S(v_{1},v_{2})$ is 1. In our case, a score closer to 1 indicated proximal subjects in the feature space. For each subject, we stored the face images of its N such proximal subjects as the potential selection pool. To generate synthetic images of existing subjects, \ie expanding a real identity, we simply created the donor pool with all the real face images of that subject.

\subsection{Attribute based Reshaping of Facial Parts}
Once the donor pool was selected, we implemented a re-shaping step before the triangle replacement. Due to variations in facial structure, positioning regions from the donor on their relative position in the base face could give the resulting synthetic face a distinctly non-real appearance (overly stretched or compressed). To address this we reshape the synthetic face using natural face shape ratios.


Previously, researchers have used ratios based on the height of the full head or hand marked forehead points in relation to perceived beauty of a face \cite{FaceAttr}. As we did not have facial landmarks of the forehead or head top points, we instead obtained our own distribution of face shape ratios from the donor set. For each ethnic group of a specific gender, we constructed a rank ordered list of the ratio of facial regions like the eyes, nose and the mouth from the subjects in the donor pool. Then we computed the inter-quartile range (IQR) from this list for each group. When synthesizing a new face for that group, the eyes, nose, and mouth regions were positioned on the target face so as to adhere as closely as possible to this IQR value. Using these face ratio ranges, the vertical positioning of the parts of the face became much more natural in appearance.


\subsection{Triangle Replacement}
After the facial reshaping, we replaced those regions with corresponding triangles from face images in the base face's donor pool. To preserve the visual uniformity of the vital facial regions, such as the mouth, nose and the eyes, we designated all triangles in that region to come from the same donor. Without this restriction, the mouth, nose and eyes appear as an amalgamation of multiple donors. This was not a concern for the rest of the face, however, so the triangles in the cheek and jaw area were chosen separately from the donor pool. Therefore, the number of donor face images used for synthesis, $C_{donor}$, can be tuned by the user depending on the faces available in the donor pool and the degree of distinctiveness sought in the synthetic face. A smaller value of $C_{donor}$ ensures the uniformity of each facial part, while a higher value makes the synthetic face appear more distinctive. For our experiments and donor pool, we found the synthetic faces looked more realistic when $C_{donor}$ was set between 7 (for smaller donor pool size) and 10.

\subsection{Adjusting Color Distribution of Triangles}
After the donor triangles were selected, their color distributions were individually adjusted to be closer to that of the base face triangles to deal with intensity changes across the face due to lighting. This was done by simply shifting the color distribution of the donor triangle to have the same mean as the base triangle that it replaced. At first, the difference in mean intensities between the base and the corresponding donor triangles was computed for each color channel and this value was used as an offset to the intensities of the corresponding channels in the donor triangle.

Without this adjustment, the synthetic face can appear noticeably splotchy in places. Adjustment in the HSV color space tends to leave a pinkish tint to some faces and does not perform well for darker skin tones. For this reason SREFI currently uses color adjustment in RGB color space.

\subsection{Blending Triangles}
Placing the triangles together on reshaped facial parts does make the synthetic face look different from the base face, but it seems unnatural as the transition of intensity across the donor triangles may not be smooth. This un-blended synthetic face can be seen in Fig \ref{fig:SREFIWorkFlow}.e. To make the synthetic face more natural, the last step of our method blends the triangles together using Laplacian pyramids \cite{LapPyr}.

The blending process starts with the base face, a mask to specify the position of the triangle on the face to be replaced, and the donor face, after reshaping and color adjustment. Then, Gaussian pyramids are created for each color channel of these three images. The images at each level of the pyramid were scaled down by a factor of 4 from the level below it by applying a Gaussian blur using a $5\times5$ kernel. Each pyramid was built with 4 such levels. We created two Laplacian pyramids from the base face and donor face Gaussian pyramids using the intensity difference between the image at a level of the Gaussian pyramid and the expanded version (by a factor of 4) of the image at the level immediately above it. The image at the top-most level of the Gaussian pyramid was stored as it is at the top-most level of the corresponding Laplacian pyramid.

A third Laplacian pyramid was generated for the blended face using the triangular mask values as a switch. For each level of the Gaussian pyramid of the mask, we added pixel values from that level of the base Laplacian pyramid if the mask value was 1, or from that level of the donor pyramid if the mask value was 0, to the third Laplacian pyramid. This was done as follows:-
\vspace{-0.1cm}
\begin{equation}
\begin{split}
    p1 = (g_{mask})_{i}*(l_{base})_{i} \\
    p2 = (1 - g_{mask})_{i}*(l_{donor})_{i}
\end{split}
\end{equation}

where $(g_{mask})_{i}$ is the image at the \emph{i}-th level of the mask Gaussian pyramid and $(l_{base})_{i}$ and $(l_{donor})_{i}$ are the images at the \emph{i}-th level of the base and donor Laplacian pyramids respectively. The image at the \emph{i}-th level of the new Laplacian pyramid for the blended face, $(l_{blend})_{i}$, was generated by simply adding the two images, \emph{p1} and \emph{p2}, together.



To integrate the blended images of different resolutions in the new Laplacian pyramid, we collapsed them together from top-to-bottom. This was done by adding the expanded version of the image at the \emph{i}-th level $expand((lap_{blend})_{i})$ and the image at the (\emph{i-1})-th level \ie $(lap_{blend})_{i-1}$. The blended image was further normalized with pixel intensities less than 0 changed to 0 and those greater than 255 changed to 255. We merged the collapsed images for the three color channels together to get the final blended image. The blended image looks quite natural as shown in Fig \ref{fig:SREFIWorkFlow}.f. Interestingly, this method achieves better regional blending for human faces than the popular Poisson blending (seamless blending) \cite{PoissonBlending}, which tries to incorporate pixels from both the source and destination in the blended region, generating faces with overlapping second nose or mouth.

\section{Experiments and Results}
To assess the realism, uniqueness and stability of the synthetic faces generated using our method, we performed an extensive set of experiments, described in this section.

\begin{figure*}
\begin{center}
   \includegraphics[width=1.0\linewidth]{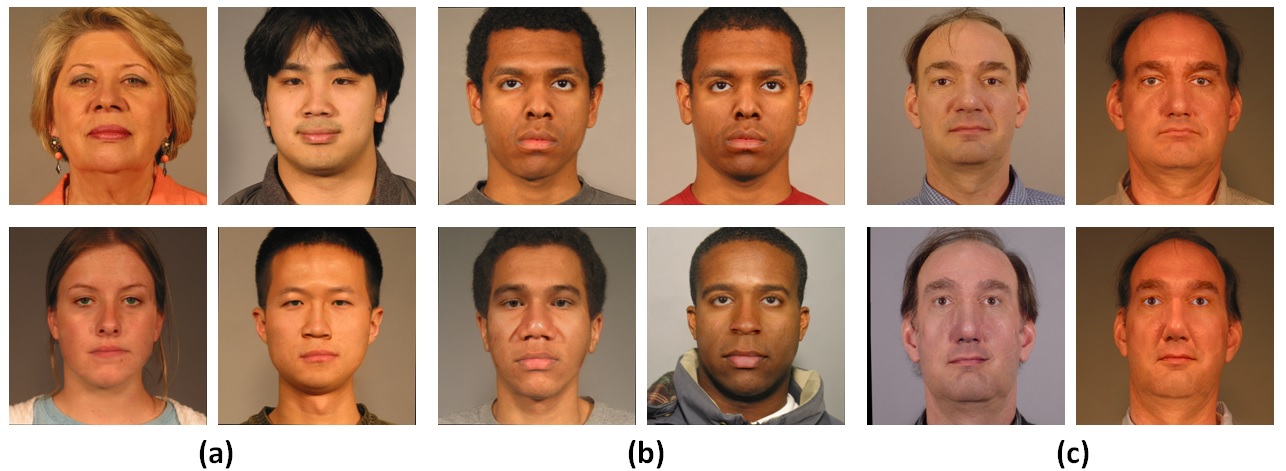}
\end{center}
   \caption{Sample face images used in the three sessions of the human study respectively - (a) TOP: two real face images, BOTTOM: two synthetic face images; (b) TOP: true match pair, BOTTOM: non-match pair; (c) TOP: real and synthetic face images of an existing identity, BOTTOM: two synthetic face images of the same real identity.}
\label{fig:HS_Imgs}
\end{figure*}

\subsection{Human Rater Study}
One way we used to assess the realism of the synthetic faces was a human rater study. The experiment had 20 novice raters who had not previously seen our synthetic faces. Each rater participated in three different experiments. In one they rated the realism of a face image, and in two other experiments they rated whether two face images came from the same subject. The experiments used the PsychoPy framework \cite{PsychoPy} to present stimuli and record the responses. The data was collected from the human raters under a IRB-approved human subjects protocol.

{\bf Experiment 1}: Raters were shown either a real or a synthetic face image, and asked to rate whether it is a real face, using a three-valued Likert scale \cite{Likert}. We selected 100 real and 100 synthetic frontally posed, 2D-aligned face images ($512\times512$ in size), 50 male and 50 female each, belonging to 3 ethnic groups (Caucasian, Asian and African American). Some example images can be seen in Fig \ref{fig:HS_Imgs}.a. Before the experiments began, raters were shown two practice trials - a real face image labeled as real and a synthetic face image labeled as synthetic. The 200 images were shown in a random order for two seconds each (however the next image appeared only after the rater had registered their response), as in \cite{FacialPerception}, and asked the question - ``Is this face image real?". The rater had to respond by pressing a key for ``Yes", ``No" and ``Cannot decide". The raters performed well in this experiment, marking the face images correctly 92\% of the time on average. The scores also suggest that they had less difficulty in detecting synthetic faces in female subjects than male subjects. Furthermore, we found the raters to be slightly more inclined towards marking a real face synthetic than a synthetic face real.
{\bf Experiment 2}: The goal of this experiment was to evaluate how reliably a pair of images of the same (different) synthetic identity are rated by a human observer as being of the same (different) person. For this experiment, 200 pairs of frontally posed and 2D-aligned synthetic face images ($512\times512$ in size) were generated. For the authentic-pair trials, one pair of images was generated for each of 50 synthetic male identities and 50 synthetic female identities. For the impostor-pair trials, 50 different-identity male image pairs and 50 different-identity female image pairs were generated. All different-identity pairs were of the same ethnicity. Before the actual trials, raters were shown two examples, one with a pair of images from the same synthetic identity, labeled as such, and a second pair from two different identities, labeled as such. The 200 image pairs were shown in a random order for two seconds each. For each pair, the rater responded to the question - ``Are these two images of the same person?" by pressing a key for ``Yes", ``No" or ``Cannot decide", similar in theme to the study described in \cite{NixonPAMI}. The human ratings from this experiment, with 82\% average accuracy, suggests that raters found it relatively harder to correctly match synthetic images than the simpler detection task. Additionally, we found them to be slightly more inclined towards marking a true match pair a non-match than vice-versa.

\begin{figure}[t]
\begin{center}
   \includegraphics[width=0.9\linewidth]{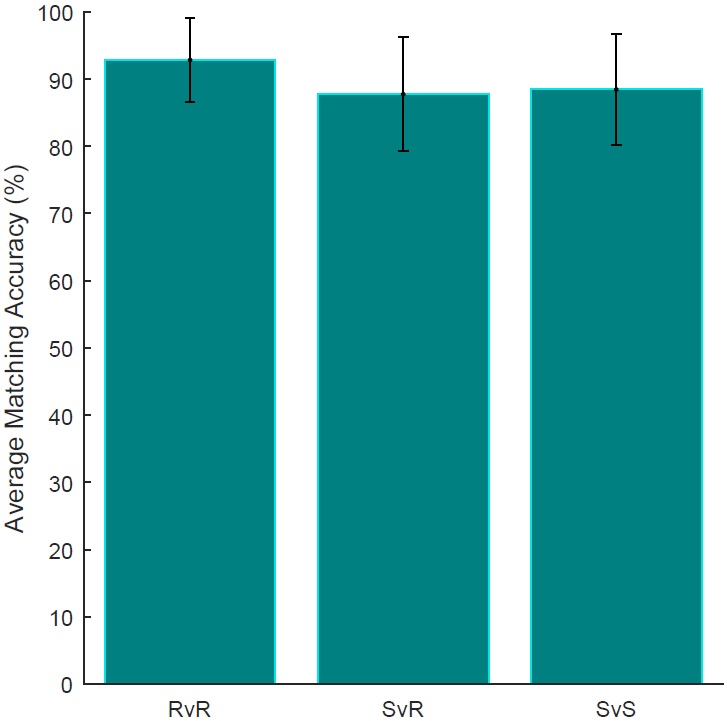}
\end{center}
   \caption{Average matching accuracy (std. dev indicated with error bars) of session 3 participants for each pair type - real vs real (RvR), synthetic vs real (SvR) and synthetic vs synthetic (SvS). }
\label{fig:HS3_Bars}
\end{figure}

{\bf Experiment 3}: The goal of this experiment was to evaluate the degree to which synthetic images of a real identity are interchangeable with real images of the same identity in a matching context. Three sets of 100 face image pairs were used - one pair of real images of each of 100 real identities, one (real image, synthetic image) pair of each of 100 real identities, and one synthetic image pair of each of 100 identities. In all three cases, the 100 pairs were split evenly between male and female. The real image pairs came from different day acquisitions. The 300 image pairs were shown in a random order for two seconds each. For each trial, the image rater answered the question - ``Are these two images of the same person?" by pressing a key for ``Yes", ``No" or ``Cannot decide". Before starting the experiments, the raters were shown three practice trials, one corresponding to each of the three conditions. The raters did better in this experiment, with the average accuracy of 90\%. However, they tended to make incorrect decisions more frequently for female face pairs compared to male face pairs. We also calculated the average matching accuracy of the 20 raters in matching a pair of real images (RvR), a synthetic image to a real one (SvR) and a pair of synthetic images (SvS) for the same subject for male and female subjects separately, as shown in Fig \ref{fig:HS3_Bars}. The error bars suggest that there is no discernable difference in performance between the three pair types. Thus, we can interchange a real image with a synthetic image of the same identity for face pair matching without any significant drop in recognition performance.

\subsection{Evaluating Uniqueness of Synthetic Faces}
To evaluate the uniqueness of synthetic face images and identities generated using our method, we performed face matching experiments using VGG-FACE \cite{VGG} pre-trained on 2.6M face images. We prepared two augmented versions of the SREFI donor set (Table \ref{Tab:SREFIData}), which had 15,939 real face images of 1471 real subjects (Real ID). The first augmented dataset contained 15,939 synthetic face images of the same 1471 real subjects as in the SREFI donor set (\ie we artificially expanded each identity (Expand ID)). The second augmented set comprised of 31,878 synthetic images of 2942 synthetic identities generated by selectively recombining elements from the SREFI donor pool (Synth ID).

\begin{figure}[t]
\begin{center}
   \includegraphics[width=1.0\linewidth]{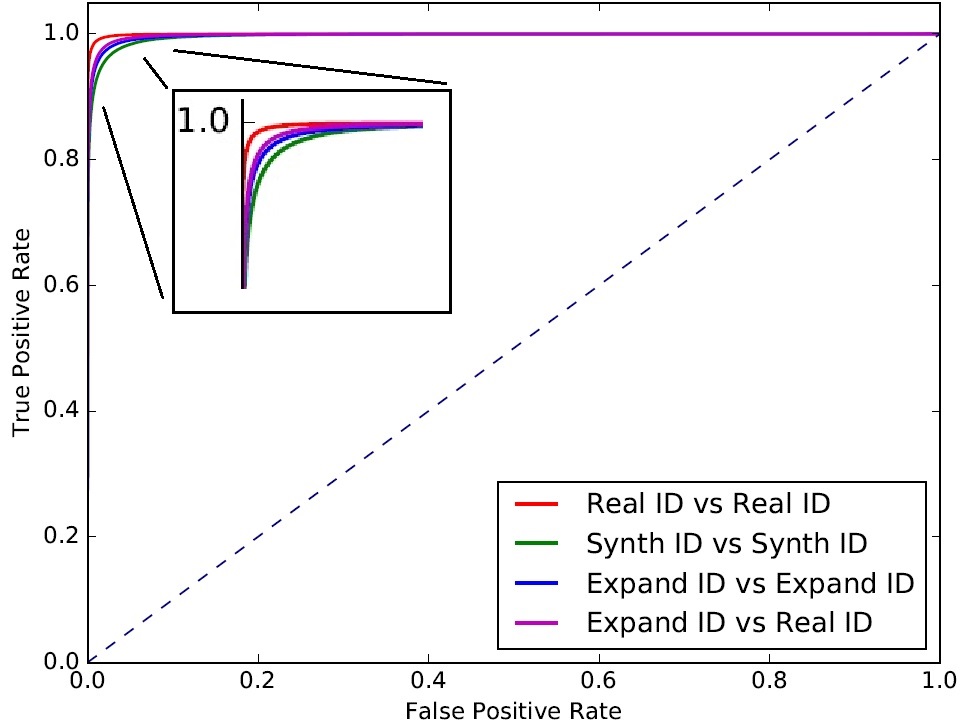}
\end{center}
   \caption{ROC curves for the four matching experiments performed with pre-trained VGG-FACE \cite{VGG} and cosine similarity.}
\label{fig:ROC_Matching}
\end{figure}

The images of the three datasets (Real ID, Expand ID and Synth ID) were supplied to the pre-trained VGG-FACE. The 4096-D vector from its fc7 layer was stored for each image as its feature representation. We performed four independent matching experiments, using cosine similarity as our scoring scheme, with these feature representations - 1) Real ID with Real ID, 2) Synth ID with Synth ID, 3) Expand ID with Expand ID, and 4) Expand ID with Real ID. The ROC curves generated for the experiments can be seen in Fig \ref{fig:ROC_Matching}. Although the accuracy while matching only real subjects is the highest, the augmented dataset generated from the real subjects matched well with each other and the real dataset as well.

The similarity in verification performance for all the three datasets suggests that the synthetic images can be used to supplement existing face image datasets by not only increasing the number of images per subject but also generating different face images of entirely new subjects without any significant loss in recognition performance. Therefore, this augmentation process can aid researchers looking to augment face datasets modest in size.

\subsection{Evaluating Stability in CNN Training}
To test the stability of the synthetic faces generated using our method for CNN training, we prepared two augmented datasets from the original SREFI donor set (Table \ref{Tab:SREFIData}). We trained networks with the VGG-FACE architecture from scratch on these datasets and used the trained CNNs to match extremely challenging face image pairs from the `Ugly' partition of the ``Good, Bad and Ugly" dataset (GBU) \cite{GBU}. Since some subjects were present in both the SREFI donor set and GBU, we removed the face images of these common subjects from the SREFI donor set. This reduced the donor pool to 10,692 face images from 1296 real subjects.

The first augmented dataset had 10,692 real images of 1296 real subjects, 10,692 synthetic images of the same 1296 subjects and 84,636 synthetic images of 10,440 synthetic subjects generated from the SREFI donor set. The second augmented dataset contained 10,692 real and 21,384 synthetic images of the 1296 original subjects and 176,098 synthetic face images from 21,538 synthetic subjects. Therefore, the two augmented datasets contained 106,020 and 208,174 face images respectively. To gauge the effectiveness of these augmented datasets, we prepared a third dataset by selecting 260,882 (real) face images from the CASIA-WebFace (CW) dataset \cite{CASIA}. All the images in these three datasets were 2D-aligned about their eye-centers extracted using the method in \cite{DLibPose}, cropped and resized to $224\times224$. We trained the VGG-FACE model from scratch with these three datasets independently for 50 epochs, using Caffe \cite{Caffe}. To maintain consistency, we used the same set of hyper-parameters, learning function (SGD) and the same NVIDIA Titan X GPU for all three training sessions.

\begin{figure}[t]
\begin{center}
   \includegraphics[width=1.0\linewidth]{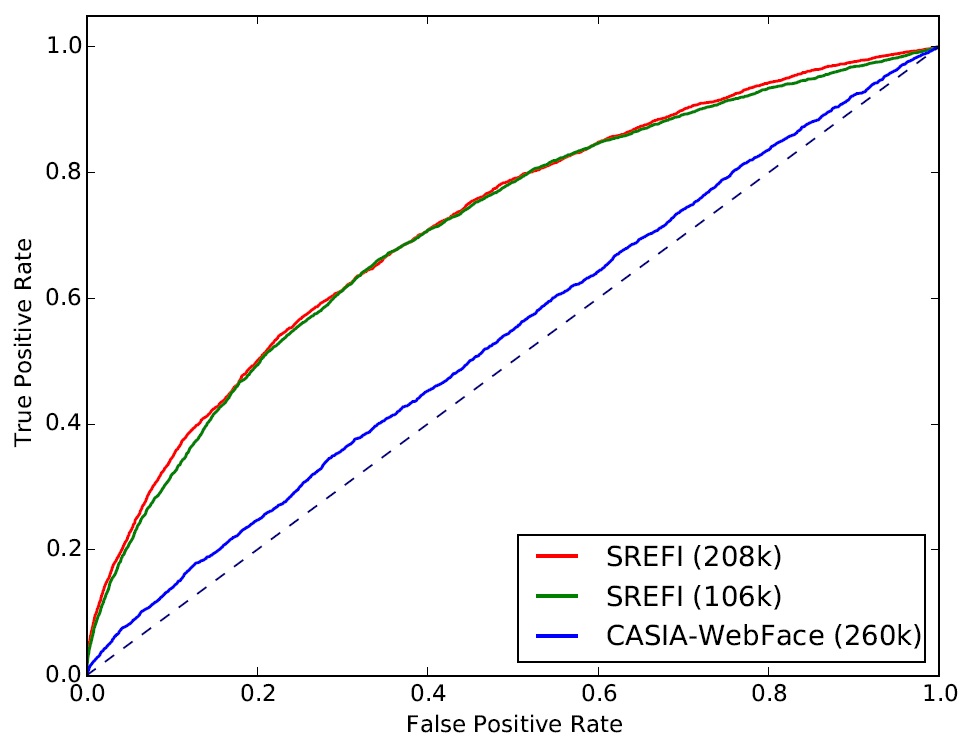}
\end{center}
   \caption{ROC curves for the matching experiments performed by training VGG-FACE \cite{VGG} with the two augmented datasets generated from our method and a subset of the CASIA-WebFace \cite{CASIA} dataset.}
\label{fig:ROC_Matching_GBU}
\end{figure}

Once training for all three sessions terminated, we used the saved snapshots to extract 4096-D feature representations from the fc7 layer of the CNN for each of the images in face image pairs from the `Ugly' partition of GBU. We normalized the feature vectors between a range of 0 and 1 using linear min-max normalization. After normalization, we computed matching scores, using cosine similarity, between feature vectors of the query and target face images in the GBU `Ugly' partition. The ROC curves generated from the three matching experiments can be seen in Fig \ref{fig:ROC_Matching_GBU}.

Interestingly, the networks trained with the two augmented datasets significantly outperformed the snapshot trained with the face images from CW. This might be due to the fact that face images in CW vary in facial pose and illumination while the original and synthetic images in our augmented datasets and the GBU face images were frontally posed. Moreover, the CNN trained with the larger augmented dataset (208k) performed better than the CNN trained with the smaller dataset (106k). This suggests that increasing the number of synthetic face images and using the augmented dataset for CNN training might further boost the network's performance in recognition experiments.

\section{Conclusion}
In this paper, we proposed a novel method for generating natural looking synthetic face images for real and synthetic subjects. This method can benefit face biometric research by - (a) replacing the manual collection of face image datasets which require abundant time and resource, and (b) augment existing face image datasets to create larger and deeper supersets for training CNN models and improve their face representation capability.

To test the fidelity of synthetic face images and identities generated with our method, we performed a human evaluation study where human raters - (1) correctly detected real and synthetic face images once shown an example, (2) correctly matched a pair of face images from the same person (real or synthetic). To test the uniqueness of these synthetic faces, we used the pre-trained VGG-FACE model's \cite{VGG} feature representations. The ROC curves generated for matching a synthetic image pair or a real and a synthetic image pair was found to be proximal to the ROC curve for matching a pair of real face images. Finally, we trained the VGG-FACE model from scratch with two augmented datasets (208k images and 106k images) generated using our method and a subset of the CASIA-WebFace dataset (260k images) \cite{CASIA} independently. We found the CNN trained with the larger dataset to outperform the CNNs trained on the smaller dataset and the CASIA-WebFace subset while recognizing face image pairs from the `Ugly' partition in the GBU dataset \cite{GBU}. This suggests the synthetic face images and identities generated using our method are stable for CNN training and can boost its face recognition performance.

As possible extensions to our work, we would like to evolve our method to generate synthetic images with various facial poses using faces extracted from video frames in the donor set along with still images. We also want to test the recognition performance of a CNN trained with a dataset containing millions of synthetic face images, generated using our method, comparable in size to the VGG-FACE dataset \cite{VGG}.

\end{document}